\documentclass[twocolumn, 10pt,letterpaper]{article}

\usepackage{cvpr}
\usepackage{times}
\usepackage{epsfig}
\usepackage{graphicx}
\usepackage{amsmath}
\usepackage{amssymb}
\usepackage{geometry}
\usepackage{tabularx}
\usepackage{multicol,lipsum}
\usepackage[ruled,vlined]{algorithm2e}

\usepackage[pagebackref=true,breaklinks=true,letterpaper=true,colorlinks,bookmarks=false]{hyperref}

\cvprfinalcopy 

\def\cvprPaperID{1234} 

\ifcvprfinal\fi
\begin{document}

\title{Cross-domain Few-shot Learning with Unlabelled Data}

\author{Fupin Yao\\
iFlyTek-Surrey Joint Research Centre on Artificial Intelligence, UK
\\
{\tt\small fupin.yao@surrey.ac.uk}
\and
}

\maketitle

\begin{abstract}
   Few shot learning aims to solve the data scarcity problem. If there is a domain shift between the test set and the training set, their performance will decrease a lot. This setting is called Cross-domain few-shot learning. However, this is very challenging because the target domain is unseen during training. Thus we propose a new setting some unlabelled data from the target domain is provided, which can bridge the gap between the source domain and the target domain. A benchmark for this setting is constructed using DomainNet \cite{peng2018oment}. We come up with a self-supervised learning method to fully utilize the knowledge in the labeled training set and the unlabelled set. Extensive experiments show that our methods outperforms several baseline methods by a large margin. We also carefully design an episodic training pipeline which yields a significant performance boost.
\end{abstract}

\section{Introduction}
Suppose you are given a project: given one labeled X-ray image for each kind of several diseases, you are asked to diagnose diseases using X-ray images. How will you solve this problem? Of course, you boss won't be pleased if you just say this is impossible. One solution is to utilize other labeled datasets available, such as ImageNet or other medical images. Then you may try domain adaptation or transfer learning methods. But wait, they will not work because you can only touch very few images in the testing set. Domain adaptation and transfer learning require a large amount of data otherwise the models will overfit the very small dataset soon. Cross-domain few-shot learning aims to solve this problem. However since the target domain is unseen during training, you are given access to some unlabelled X-ray images of diseases which are different from the diseases in the test set. This problem is called cross-domain few-shot learning with unlabelled data. we will discuss it in detail in this paper below.

\begin{figure}[t]
\centering
	\includegraphics[scale=0.37]{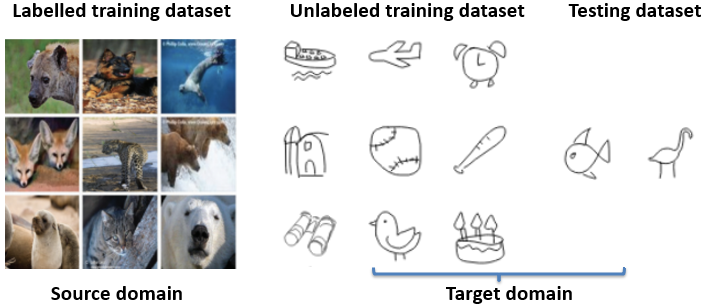}
	\caption{We have 3 datasets in this problem: a labelled training set from the source domain, an unlabelled training set from the target domain and a testing set from the target domain \cite{guo2019} \cite{QuickDra80:online}}
\end{figure}

Deep learning become extremely successfully in the last decade. The image recognition performance on ImageNet \cite{imagenet_cvpr09} has surpassed the performance of humans []. Google and Apple built remarkable intelligent speech assistant using speech recognition and speech synthesis technology. Self-driving cars has been invested a lot and given high expectations. We can see them in the near future if we are lucky. However, most successful deep learning systems are supervised learning systems which means that they demand a large a mount of labelled data. The consequence is that it will become extremely difficult if you want to train your model on mobile devices and you can't expect your model to work well if labelling is too expensive.

Few-shot learning is proposed to alleviate the "big data demand" problem in deep learning and has achieved a lot during the last several years. In the standard few-shot learning benchmark, there are only 1 or 5 images in each class, and they are called one-shot learning and 5-shot learning respectively. Most researchers address this few-shot problem using meta-learning. Meta-learning, also known as learning to learn, aims to learn a model which learn new skills or adapt to new environments rapidly \cite{weng2018metalearning}. The training set of meta-learning includes lots of episodes each of which is a few shot learning task. Lots of meta-learning algorithms \cite{snell2017rototypical} \cite{finn2017odelagnostic} \cite{lee2019etalearning} \cite{chen2019} has been proposed and get better and better performance on several benchmarks. However, when the test set comes from a different distribution, the performance of most meta-learning algorithms will degrade a lot, which is first noticed by \cite{chen2019}. In addition, the case where the test set comes from another distribution happens a lot in the real world. When the model is deployed, we will always see the testing samples vary greatly from the training distribution. 

Cross-domain few-shot learning is dedicated to solve the above distribution shift problem. There are several papers in this topic, and \cite{tseng2020rossdomain} \cite{guo2019} are representatives of them. However, they assume there are several source domains, which is a cumbersome requirement. And, with only 3-5 domains, there is little hope that the model can generalize to another unseen domain. Most importantly, it assumes you can't touch the target domain which is not a valid assumption and makes the problem extremely difficult to solve.

Cross-domain few-shot learning with unlabelled data is a new setting which is proposed in this paper. A unlabelled dataset from the target domain is provided to bridge the gap between the source domain and the target domain. In most of the cases, unlabelled data is easy to collect since no labels are required. In addition, the access to the target domain is not a problem in most of real life problems. After some unlabelled data is added, we have a better chance to solve the few shot learning problem in the target domain. So this is a more realistic and feasible setting. We will show our arguments by extensive experiments in later sections.

In summary, the contributions of this paper are as follows:

1. We propose a more realistic and feasible setting.

2. We construct a dataset for this setting so that algorithms under this topic can be fairly compared.

3. A novel self-supervised learning based algorithm is proposed for this problem and it's proven to be effective by extensive experiments.
\section{Related work}
\subsection{Cross-domain few-shot learning}
Cross-domain few-shot learning aims to deal with the domain shift problem in few shot learning when the testing dataset is from another domain which is different from the meta-training dataset. \cite{tseng2020rossdomain} first proposed a feature-wise transformation architecture to solve this problem. A few feature-wise transformation layers are inserted into the feature extractor to simulate feature distributions extracted from the tasks in various domains. The whole pipeline is a learning to learn approach and they hope to learn these hyper-parameters of these feature-wise transformation layers. In the meta-testing phase, they hope their models can generalize to unseen domains. \cite{chen2019} proposes a more realistic setting where the domain shift between the training domains and the testing domains are much larger. For example, the training images can be natural images while the testing images can be medical images or satellite images. They also proposed several baseline methods in the paper. Although this is a more realistic setting, we claim that it's too challenging since we can't tough the testing domains and those testing domain has a vast domain shift from those training domains. The accuracies on medical image datasets are very low.

\subsection{Heterogeneous domain adaptation}
Our problem setting is related to heterogeneous domain adaptation. Traditional domain adaptation targets the cases where the testing data distribution is different from the training domain. In heterogeneous domain adaptation (classification), the label spaces of training set and the testing set are also different. From example, in your testing datasets, you have some classes which are unseen in the training dataset, which is very common in practice as you can't expect your classifier to work in the exact same environment when you train it.

This is a less studied research topic and there are very few papers. \cite{chang2018isjoint} proposes a method using a shared space between the source and target domain. It trains the model with an unsupervised factorisation loss and a graph-based loss. \cite{gao2020deep} adopts a deep clustering method for this problem.

\subsection{Unsupervised domain adaptation for Person/object ReID}
We also find our problem is related to unsupervised domain adaptation for Person/object ReID. Researchers study the domain adaptation in person/object ReID because it is observed that domain shift is also very common in person/object re-identification where the testing images have very different characteristics from the training images. For example, there maybe more occlusions and worse illuminations and extreme weathers. 

There are two types of methods in this field, domain translation based methods,such as using \cite{zhu2017npaired} to translate images from the target domain to the source domain and then train a model with translated images. Another type of methods is pseudo-label based method in which clustering techniques are used to extract pseudo-labels for the unlabeled target data and then unlabeled data with pseudo-labels is used to train a model.

The person IDs in the target and source domain are disjoint, which is the same as our problem. But unsupervised domain adaptation for person/object ReID is targeted on person/object ReID only while are are studying a much broader problem. In addition, they don’t have class labels but have person ids which can be viewed as fine-grained class labels. 

\subsection{Self-supervised learning}
Self-supervised learning becomes really hot these years which tries use the supervision signals from the data itself instead of from labels. This reduce the cost of labelling.

\textbf{Contrastive learning} is now widely used in self-supervised learning. In contrastive learning we try to find positive and negative samples and force the model to obtain a small distance between a reference sample and its positive samples and big distance between the reference and negative sample. For example, in SimCLR (\cite{chen2020}), augmented images of the same image are treated as positive samples while other images are treated as negative samples. Using this distance supervision signal we can learn meaningful representation even though we may not have labels. Recent self-supervised methods are going to surpass the performance of supervised methods. Our baseline method is based on this type of method.

\textbf{Context based self-supervised learning} This type of self-supervised learning utilize the information from the context of the current data. For example, \cite{mikolov2013fficient} predicts missing words according to neighboring words. \cite{noroozi2016nsupervised} predicts the relative positions of image blocks. \cite{gidaris2018nsupervised} predicts the angles after images are rotated. Our method is based this rotation based supervised learning method.
\section{Cross-Domain Few-Shot Learning with unlabeled data}
\subsection{Motivation}
Without touching the target domain, getting a good performance on the unseen testing domain becomes extremely challenging. To bridge the gap between the training and testing domain, we provide some unlabeled data from the target domain during the training phase to make the problem less challenging. We have this modification to the cross-domain few-shot learning for the following reasons:

1. Unlabeled data are easy to collect as you don't need experts to label them, Even for medical X-ray images, machine produce billions of images every year and you don't need doctors to label them one by one.

2. Access to the target testing domain is easy to obtain in most cases. You can easily collect data from the same domain where the target domain is from. The only exception is the case where the data generation is very expensive. For example, if you want to collect car accident images for auto-driving cars. But these cases are very rare because the car accident rate is very low. We don't study this type of problem in this paper.

Therefore, our setting is more realistic as we have large domain shifts and also more feasible since we have access to the target testing domain while very less limitations in terms of applications.

\subsection{Problem formulation}
In our proposed setting, cross-domain few-shot learning with unlabeled data, we have 3 datasets: $D_{train}, D_{unlabelled}, D_{D_{test}}$. $D_{train}$ is the meta-training dataset from the training domain $A$, $D_{unlabelled}$ is the unlabelled dataset from the testing domain $B$ and $D_{test}$ is the testing dataset from the testing domain $B$. There three dataset also have different label spaces: $L_{train}, L_{unlabelled}, L_{test}$.

\textbf{Learning objective} Our learning objective is get good performance on few-shot tasks sampled from $D{test}$ after training models on $D_{train}$ and $D_{unlabelled}$. Each few-shot task is 5-way 5-shot or 5-way 1-shot task same as the standard few-shot learning. Here the number for shot means the number of images in each class and the number of ways means the number of classes in the task.
\section{Proposed benchmark}
\subsection{Benchmark requirements}
As we propose a new setting, we a new benchmark to compare the performance of different algorithms. We have several requirements to satisfy for our setting:

1. The domain shift between the testing dataset and the training dataset should be large enough. This is to make our setting more realistic as you will most always seen vast domain shift when your models are deployed.

2. We should have multiple different domains to choose. The effectiveness of a algorithm should not depend on the choice of domain and it should be consistent across different domains. So we should have multiple different domains.

3. The number of images and the number of classes should not be too small or large. If they are too small, the setting becomes less realistic. If they are too large, most researchers will waste lots of computing resources which is not worthwhile.

We have several choices at hand to choose. \cite{guo2019} proposed a benchmark where there is only one source domain and four other testing domains which has bigger and bigger domain shifts. The training dataset is ImageNet \cite{imagenet_cvpr09} and the testing sets are crop disease dataset \cite{mohanty2016using}, Euro satellite dataset \cite{helber2019eurosat}, medical dataset with colors \cite{tschandl2018ham10000} and medical X-ray dataset \cite{wang2017hestxray8}. We don't choose this because there is only one source dataset and some of those dataset are too big.

\cite{tseng2020rossdomain} uses CUB \cite{wah2011caltech}, Cars \cite{krause20133d}, Places \cite{zhou2017places} and Plantae \cite{van2018inaturalist}. However, several of these datasets are natural images and the domain shifts are not big enough to be more realistic. And also, several of these datasets are too big.

In the end, we choose DomainNet \cite{peng2018oment}. This is one of the largest domain adapation dataset. It has 6 domains and 345 classes for each domain. We discard 'Infograph' as it is too noisy and we can't learn any effective signals from it. We also discard 'quickdraw' because images in it contain too less information and it's too challenging. Discarding these two domain is a common practice in domain adaptation research and domain generalization \cite{jin2020eature}. In the end, we have 4 domain: real, painting, sketch and clipart. 'real' contains natural images, 'painting' are painting images, 'sketch' includes sketch images and 'clipart' is a collection of clip art images. A quick overview of DomainNet is in figure 2.

 To keep the number of classes in each split as consistent as the standard few-shot learning, we split the 345 classes as 64 classes for training, 261 classes for unlabelled dataset and 20 for the testing dataset.
 
 \begin{figure}[]
\centering
	\includegraphics[scale=0.16]{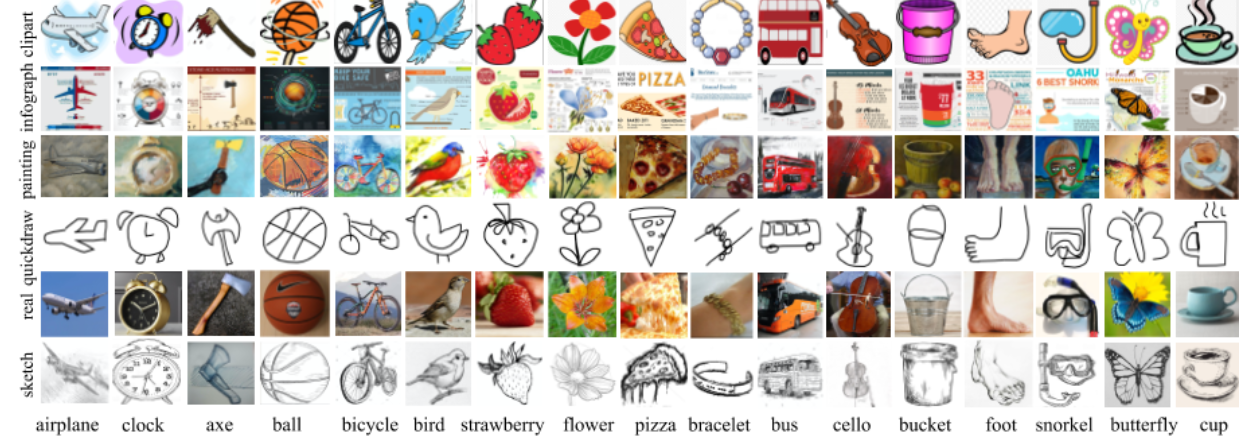}
	\caption{DomainNet overview}
\end{figure}

\section{Self-supervised learning for cross-domain few-shot learning}
\subsection{A baseline}
Before we introduce our own method, we first introduce one of the baseline we have experimented. We will use this as a baseline. The method is inspired by (\cite{ge2020elfpaced}) The original method is designed for object re-identification and we re-purpose it for our setting. First we use the clustering algorithm DBSCAN (k-means can also be applied) to cluster the unlabeled target domain data. Then we have 3 sets of samples: labeled source dataset, clustered unlabeled target dataset and unclustered samples in the unlabeled target dataset. Many methods in UDA Re-ID discard those unclustered data but we think it’s a waste of data so we propose a unified framework to fully utilize all the data we have:
\begin{equation}
    \resizebox{1.0\hsize}{!}{$L_f = - log\frac{exp(<f, z^+>/\tau)}{\sum_{k=1}^{n^s}exp(<f, w_k>/\tau)+ \sum_{k=1}^{n_c^t}exp(<f, c_k>/\tau) + \sum_{k=1}^{n_o^t}exp(<f, v_k>/\tau)}$}
\end{equation}

This is our loss function in which f is a feature vector of an image either from the source dataset, clustered unlabeled target dataset or uncluster outliers. $z^+$ indicates a positive prototype corresponding to $f$. $w_k$ is the centroid of all features in the class $k$, $c_k$ is the centroid of all features in the cluster $k$ and $v_k$ is the $k^{th}$ outlier feature. If f is from the source dataset, then $z^+=w_k$ is the centroid feature vector of class k that f belongs to; if f is from the clustered dataset, then $z^+=c_k$ is the centroid feature vector of the cluster $k$ that f belongs to and if $f$ is from the unclusted outliers then $z^+=v_k$ is the outlier feature that f belongs to.

Our algorithm alternates between updating the model using the above comparative learning loss and producing pseudo-labels using clustering methods. Detailed description is show in algorithm 1.

\begin{algorithm}
	\KwResult{Classification accuracy}
	Let N be the number of iterations for updating the model after each clustering\;
	\While{Traing}{
		Obtain pseudo-labels for clustered samples and unclustered samples using DBSCAN algorithm with the unlabeled dataset\;
		$iter$ = 0\;
		\While{$iter<N$}{
			$iter = iter + 1 $ \;
			Updating the model using the loss function in formula 1.1 with all labeled samples in the source dataset, clustered samples with their pseudo-labels and unclustered samples\;
		}
	}
	\caption{Baseline}
\end{algorithm}
\subsection{Self-supervised learning method}
Inspired by recent progress in self-supervised learning and semi-supervised learning \cite{zhai20194l}, we proposed a new self-supervised learning method for our new setting.

Rotation is a widely used technique in self-supervised learning. They first rotation these images and then predict degrees of rotation. The intuition behind this is that we should fully utilize the relative rotations among images and use them as supervision signals which will greatly help us to learn general image representation. These feature representations can be used for any downstream tasks since they are not trained on any specific tasks during self-supervised learning. All you need to do is fine-tuning these representations.

\begin{figure}[]
\centering
	\includegraphics[scale=0.6]{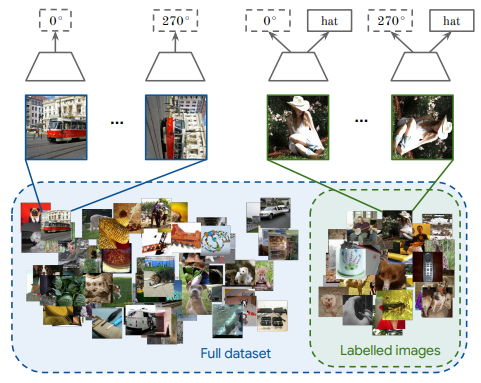}
	\caption{Method overview: our method is based on rotation-based self-supervised learning. We predict degrees for rotated images in both the labeled training set and the unlabelled dataset and we also have another supervised loss using class labels.}
\end{figure}

Our method adopts the rotation idea and works as following: we have two training phases. The first one is feature extractor training and the second one is episodic training. Two-phase training is a standard practice in few-shot learning and it performs much better compared to one-phase training.

\textbf{Phase 1: feature extractor training} We rotate each image in the labeled training set and the unlabelled dataset for 0, 90, 180 and 270 degrees and then we use these degrees as labels to form the self-supervised loss. In addition, since our labeled training set have labels, we have another supervised loss. We train our model with these two loss functions together. In this way, our model fully utilize the knowledge in both the labeled training set and unlabelled set. Since the target domain has been seen during training, we will seen a great performance boost during testing compare to cross-domain few-shot learning without unlabelled data. The overview of feature extractor training is in figure 3.

The self-supervised loss and supervised loss use the same feature extractor. There is no need to have feature extractor with different parameters. Using the feature extractor can reduce the number of parameters and make the feature representation more generalized. 

\textbf{Phase 2: episodic training} We follow the standard few-shot learning practice to do the episodic training. We sample 1000 tasks in each epoch and train the model for few hundreds of epochs. We use the mean centroid classifier, the same as Prototypical neural networks \cite{snell2017rototypical}. In this way, our model learns to deal with few shot tasks after only seeing a few training examples. 

The whole pipeline is in algorithm 2.
\begin{algorithm}
    \textbf{Phase 1: feature extractor training}\\
	Let $N_1$ be the number of epochs\;
	$iter$ = 0\;
	\While{$iter < N_1 $}{
	    Randomly sample images from the labeled set or the unlabelled dataset, rotate them and get the self-supervised loss using rotation degrees as labels\;
	    Use the above images from the labelled training set and class labels to form the supervised loss\;
		Update the model using the two above losses\;
		$iter = iter + 1 $\;
	}
	
	\textbf{Phase 2: episodic traing}
	Let $N_2$ be the number of epochs\;
	$iter$ = 0\;
	\While{$iter<N_2 $}{
	    Randomly sample 5-way 1-shot and 5-way 5-shot tasks from the labeled training set\;
	    Use the tasks and the standard mean centroid classifier to train the model\;
		$iter = iter + 1 $\;
	}
	\caption{Self-supervised learning for cross-domain few-shot learning with unlabelled data}
\end{algorithm}

\section{Experiments}
\subsection{Datasets}
As dicussed in the section 'Proposed benchmark', we use 'real' as the source domain where the training set belongs to and others as the target domains which the unlabelled dataset and the testing dataset are sampled from. In each domain, we have about 40000 images.

\subsection{Model architecture and training details}
There are four commonly used neural network architectures in standard few shot learning: Conv4 (4 convolutional layers), ResNet 12, ResNet 18 and WideResNet. The size of neural networks should not be too small as \cite{chen2019} points out that too small neural networks can't fully extract useful feature representations from images and the differences among different architectures are much larger than the differences among different few shot learning methods with the same architecture. The size can't also be too large. Otherwise they consume too much resource to train models. So Resnet 18 is a good choice and we use it for all of our experiments.

As mentioned earlier, we first train a feature extractor and then do the episodic training. During the episodic training, we sample 100 hundreds episodes for each epoch and train our model with 200 epochs. During testing, to get accuracies with lower variance, we test our models on 10000 episodes.

\subsection{Result}
We compare our method with feature extractor trained on the 'real' domain (`Backbone` in the table) and the constrative learning baseline ('Baseline' and 'Baseline1' in the table).

We can see from the table our method surpasses all the baselines on most of target datasets and the margin is also big enough. This shows that our method is very effective as we expected. Our result is shown in table 1.

\begin{table}[h!]
\begin{center}
	\begin{tabular}{ |p{1.25cm}|p{1cm}|p{1cm}|p{1cm}|p{1cm}| }
		\hline
		& clipart & painting & sketch & average \\
		\hline
		Backbone & 65.17 & 59.13 & 53.62 & 59.31\\
		\hline
		Baseline & 66.18 & 58.60 & 56.38 & 60.39\\
		\hline
		Baseline1 & 69.78 & 60.94 & 58.76 & 63.16 \\
		\hline
		Our method & 75.37 & 66.37 & 71.51 & 71.08 \\
		\hline
	\end{tabular}
\end{center}
\caption{Results of contrastive learning, 'Backbone' means feature extractors trained on the real domain without unlabeled data; 'Baseline' is the contrastive learning based approach we introduced earlier; 'Baseline1 is the result after we increased the size of the unlabelled dataset. Our method uses the increased dataset.'}
\end{table}

\subsection{Ablation study}
\textbf{The size of the unlabeled dataset} We tried to investigate the effects of the size of the unlabelled dataset as we have observed that the baseline method performs better when the unlabelled dataset is bigger. The original unlabeled target dataset has only about 2-3k images, much smaller than the source dataset 20-40 k images.

We increase the size of unlabeled dataset for two reasons: First, in most cases, unlabeled dataset are easy to build because no labels are needed and we don’t need experts to label them. Second, most unsupervised learning methods consume more samples compared to supervised learning methods. By increasing the size of the unlabeled dataset, we have more room for these algorithms. 

Our result is shown in table 1 after the size of unlabeled dataset is increased. From the table we can see that the baseline shows a big improvement on accuracies. However, our own method still outperforms this result by a large margin.
\section{Discussion and Future Work}
In this paper, we have proposed a new new setting, which is cross-domain few-shot learning with unlabelled data. We proved it's significance and importance. We hope more and more researchers to put more efforts on this important research topic.

In addition, we established a new benchmark for this new research problem. Our new benchmark is carefully designed and ideal for this problem. We hope it will be widely adopted by the research community.

We also proposed an self-supervised learning base approach to fully utilize the knowledge in the labelled training set and unlabelled dataset and bridge the gap between the source domain and the target domain. Our experiments show that our method is super effective and beat the baseline by a large margin.

While our method shows promising results,there is still large room to further investigate the problem. For example, how much can the best self-supervised learning algorithms help this problem? Does self-supervised learning really reduce the domain gap or just learns more useful and generalized knowledge?

Nevertheless, we hope that this work inspires other researchers in the field to continue exploring this problem.
{\small
\bibliographystyle{ieee_fullname}
\bibliography{egpaper}
}

\end{document}